# A Novel Framework for Learning Stochastic Representations for Sequence Generation and Recognition

Jungsik Hwang, Ahmadreza Ahmadi

*Abstract*—**The ability to generate and recognize sequential data is fundamental for autonomous systems operating in dynamic environments. Inspired by the key principles of the brain—predictive coding and the Bayesian brain—we propose a novel stochastic Recurrent Neural Network with Parametric Biases (RNNPB). The proposed model incorporates stochasticity into the latent space using the reparameterization trick used in variational autoencoders. This approach enables the model to learn probabilistic representations of multidimensional sequences, capturing uncertainty and enhancing robustness against overfitting. We tested the proposed model on a robotic motion dataset to assess its performance in generating and recognizing temporal patterns. The experimental results showed that the stochastic RNNPB model outperformed its deterministic counterpart in generating and recognizing motion sequences. The results highlighted the proposed model's capability to quantify and adjust uncertainty during both learning and inference. The stochasticity resulted in a continuous latent space representation, facilitating stable motion generation and enhanced generalization when recognizing novel sequences. Our approach provides a biologically inspired framework for modeling temporal patterns and advances the development of robust and adaptable systems in artificial intelligence and robotics.**

## I. INTRODUCTION

PREDICTIVE coding [1], [2] and the Bayesian brain hypothesis [3], [4] offer a framework for understanding how the brain processes sensory information. One of the key ideas in predictive coding is that the brain continuously generates predictions about incoming sensory stimuli and minimizes prediction errors by updating beliefs [1], [2], [5]. The Bayesian brain suggests that the brain operates as a probabilistic inference machine, constantly updating its beliefs about the world through Bayesian inference. It posits that the brain combines prior knowledge with new sensory evidence to form posterior beliefs. This approach enables the brain to process information efficiently, handle uncertainty, and adapt to new situations by updating its internal representations [3], [4].

By modeling the cognitive processes of the brain, artificial agents can improve their sensory perception, motor control, and adaptability [6], [7], [8], [9]. Recently, several deep learning models based on predictive coding and the Bayesian brain have been proposed in the fields of artificial intelligence and cognitive robotics, such as PredNet [10], Predictive Coding Network (PCN) [11], Predictive Coding Visuo-Motor Dynamic Neural Network (P-VMDNN) [12], Predictive-Coding inspired Variational RNN (PV-RNN) [13], and Predictive Coding Recurrent Neural Network (PC-RNN) [14]. These models have been used for various tasks in machine learning and robotics.

In line with these studies, we propose a stochastic neural network for modeling multidimensional sequences. The proposed model is based on predictive coding and the Bayesian brain hypothesis and it builds upon two previous models: the Recurrent Neural Network with Parametric Biases (RNNPB) [15] and the Variational Autoencoder (VAE) [16]. The RNNPB model and its variants have been widely used in cognitive robotics for modeling sequential data, leveraging the ability to learn representations in the parametric bias (PB). The RNNPB model uses the same network for both the generation and recognition of actions by sharing internal neural representations. This makes the RNNPB model ideal for tasks requiring both the generation and recognition of sequential data [17], [18].

In this study, we introduce stochasticity into the PB by adopting the reparameterization trick used in VAEs [16]. Previous studies [3], [4] emphasize the importance of probabilistic approaches in understanding neural coding and brain function. In addition, many studies have shown that introducing stochasticity can have practical benefits, such as improved generalization to unseen data and quantifying uncertainty for tasks with noisy data [19], [20], [21], [22]. By introducing stochasticity into the PB, we aim to enhance the model in these aspects, including capturing the underlying distributions of the data, reducing the risk of overfitting, and handling uncertainty in sequence generation and recognition tasks.

We validate the performance of our model on a robotic motion dataset in the context of generating robot motions and recognizing the robot's own motions. These capabilities are essential for social and collaborative robots that interact physically with humans [23]. During training, the proposed model is optimized for generative modeling of motion sequences by learning its parameters, including the PB. After

Jungsik Hwang and Ahmadreza Ahmadi were previously affiliated with the Okinawa Institute of Science and Technology (OIST). Currently, Jungsik Hwang is with Samsung Research, Seoul, Republic of Korea, and Ahmadreza Ahmadi is with Geobotica, Queensland, Australia. Correspondence should be addressed to Jungsik Hwang (e-mail: jungsik.hwang@gmail.com).

The code is available at https://github.com/mulkkyul/stochasticRNNPB.

training, the proposed model can perform inference through two pathways: it generates motions in an auto-regressive manner from its internal beliefs (top-down prediction), and it refines the predictions by minimizing discrepancies with sensory inputs (bottom-up recognition). This approach enables the robot to adapt its internal stochastic representations, ensuring stable motion generation and accurate self-motion recognition.

In summary, our contributions are threefold. First, we propose a novel stochastic model based on integrating RNNPB and VAE. This integration enables the model to perform approximate Bayesian inference and handle uncertainty, resembling the core functions of the brain. Second, we demonstrate through experiments that the stochastic model improves generalization, robustness, and adaptability in generating and recognizing motion sequences compared to deterministic models. Third, we establish connections between our model and theoretical frameworks such as predictive coding and the Bayesian brain hypothesis. This alignment between the machine learning model and biological principles offers an opportunity to investigate how cognitive processes might be implemented in the brain [6], [7], [8], [9].

## II. Related Works

### A. Neural Network Models

The proposed model builds upon two artificial neural network models: the Recurrent Neural Network with Parametric Bias (RNNPB) and the Variational Autoencoder (VAE). By integrating key characteristics from both models, we aim to enhance their capabilities and address their limitations.

The RNNPB model [15] is designed to learn and represent temporal patterns by encoding them into the parametric bias (PB). The PB captures the underlying structure and key features of the sequences, enabling the model to perform both generation and recognition of sequences. An interesting aspect of RNNPB is that similar regions in the latent space of the PB are activated during both the execution and observation of similar actions. This property allows the model to tightly intertwine generation and recognition as in the brain [17].

RNNPB and its variations have been widely used in cognitive and developmental robotics. In [15] and [17], RNNPB was used to imitate and generate human-like actions by learning combinatorial action sequences. More recently, Hwang and Tani [24] investigated the generation of creative robot motions using RNNPB. The results showed that employing a different learning method resulted in a different landscape of the PB and varied levels of creativity. The study highlighted that the latent space of PB played a crucial role in creating diverse and novel motions.

While these deterministic RNNPB models can generate and recognize sequences based on learned PB values, they operate on specific point estimates rather than data distributions. As a result, they cannot model the uncertainty inherent in data. This contrasts with other generative models like VAEs which aim to capture the data distribution. Moreover, it has been known that deterministic models are generally more prone to overfitting than stochastic models [19]. In short, the deterministic nature of RNNPB can limit its flexibility and ability to capture the full variability of the data.

The VAE introduced in [16] is a generative model that learns probabilistic latent representations of data. A VAE consists of an encoder, which maps input data to a stochastic latent space, and a decoder, which generates data from these latent variables. By jointly training the encoder and decoder to reconstruct the input data, the model learns meaningful stochastic latent representations for generation.

VAEs have been widely used in various applications, such as computer vision [25], natural language processing [26], and sequence modeling [27], [28]. Several studies have investigated the development of the latent space in VAEs. In [29], the authors introduced $\beta$-VAE, a variation of the VAE that promotes disentangled latent representations. By introducing a weighting factor $\beta$ in the VAE objective, the model is encouraged to learn independent factors of variation in the data. Similarly, Burgess et al. [30] showed that the weighting term in the loss function of $\beta$-VAE plays an essential role in shaping the latent space of the models.

VAEs, in general, are used for generative tasks rather than recognition (posterior estimation) of novel data. In VAEs, posterior estimation is implemented as a feedforward computation in the encoder network using the observation as input. This process differs from a fundamental principle in the brain, as it does not involve an interplay between feedforward and feedback processes for minimizing prediction error [1], [2]. Moreover, the feedforward computation of posterior estimation contrasts with the Bayesian brain perspective which considers perception as an optimization that combines sensory input with prior expectations [5]. In short, while VAEs provide a computational framework for representing internal beliefs as probability distributions, they lack iterative inference mechanisms involving feedback, making them different from brain-inspired models.

### B. Theoretical Frameworks

Predictive coding [1], [2] suggests that the brain continuously generates predictions about incoming sensory inputs and updates its internal models based on prediction errors—the differences between predicted and actual inputs. This iterative process minimizes surprise by refining predictions over time. The Bayesian brain posits that the brain performs probabilistic inference, representing information in terms of probability distributions and updating beliefs according to Bayesian principles [3], [4]. Under this hypothesis, the brain combines prior knowledge with sensory evidence to form posterior beliefs, enabling optimal decision-making under uncertainty.

Recently, several studies have introduced deep learning models based on these theoretical frameworks. For instance, Lotter et al. [10] introduced PredNet, a predictive coding-inspired neural network for video prediction. PredNet models the brain's ability to predict future sensory inputs by minimizing the prediction error between predicted and actual frames in video sequences. In [11], a deep predictive coding network (PCN) for object recognition was introduced. It aims to capture both feedforward and feedback information processing to improve recognition accuracy. The Predictive Coding Recurrent Neural Network (PC-RNN) [14] extends this concept to temporal sequences by incorporating recurrent connections. Experiments show enhanced performance in sequence prediction and continual learning scenarios. In [12], [31], the authors introduced a Predictive Visuomotor Deep Dynamic Neural Network (P-VMDNN) for multimodal learning in robots. They demonstrated that by minimizing the prediction error of one modality, the model could forecast the future sensory input of another modality. Ahmadi and Tani [13] introduced the Predictive Coding-inspired Variational RNN (PV-RNN) for online prediction and recognition. The model integrates variational inference with recurrent dynamics to handle temporal sequences. Experiments showed that PV-RNN effectively predicts future inputs and recognizes patterns in sequential data. Several studies have also incorporated the Bayesian brain hypothesis into neural network models. For instance, Blundell et al. [19] introduced Bayes by Backprop for performing variational Bayesian inference in neural networks. Gal and Ghahramani [20] showed that dropout in neural networks can be interpreted as a Bayesian approximation method. They demonstrated that applying dropout during test time allows for estimating model uncertainty. See [32] and [33] for a review of these approaches.

In this study, we propose a novel stochastic neural network model rooted in these theoretical frameworks. Incorporating stochasticity into the proposed model reflects the probabilistic reasoning posited in the Bayesian brain hypothesis. By updating the model's internal belief through prediction error minimization during recognition, our model aligns closely with the predictive coding principles of the brain. The proposed model provides a simple yet robust framework by combining the deterministic sequence learning capabilities of RNNPB with the probabilistic generative modeling of a VAE. While models like PCN and PC-RNN focus on predictive coding principles, they do not explicitly incorporate a probabilistic generative mechanism akin to VAEs. By integrating such a mechanism, our model aims to enhance the ability to capture data distributions and model uncertainty more effectively. Additionally, the proposed model performs variational inference in the PB, resulting in sequence-level uncertainty quantification. This approach may offer advantages in simplicity and computational efficiency compared to methods that model uncertainty at the level of outputs, hidden neurons, or weights of the neural network models.

## III. Proposed Neural Network Model

In this section, we present the stochastic RNNPB model, which is designed to learn stochastic representations of multidimensional sequences.

### *A. Model Architecture*

Fig. 1 illustrates the architecture of the proposed model, which consists of the stochastic PB layer, the input layer, the LSTM layer, and the output layer. The key innovation of the proposed model is the introduction of stochasticity in the PB. In contrast to the deterministic PB of previous studies [15], [17], our approach employs variational inference to learn a probability distribution over the PB. Specifically, the model learns the parameters of a Gaussian distribution ($\boldsymbol{\mu}$ and $\boldsymbol{\sigma}$) for each sequence in the training data. This is achieved by applying the reparameterization trick (1) to the PB, similar to the approach in VAE [16]. It allows gradients to flow during backpropagation, enabling the model to learn the distribution parameters effectively.

$$\boldsymbol{PB}^{(i)} = \boldsymbol{\mu}^{(i)} + \boldsymbol{\sigma}^{(i)} \odot \boldsymbol{\varepsilon} \qquad where\ \boldsymbol{\varepsilon} \sim N(\boldsymbol{0}, \boldsymbol{I}) \tag{1}$$

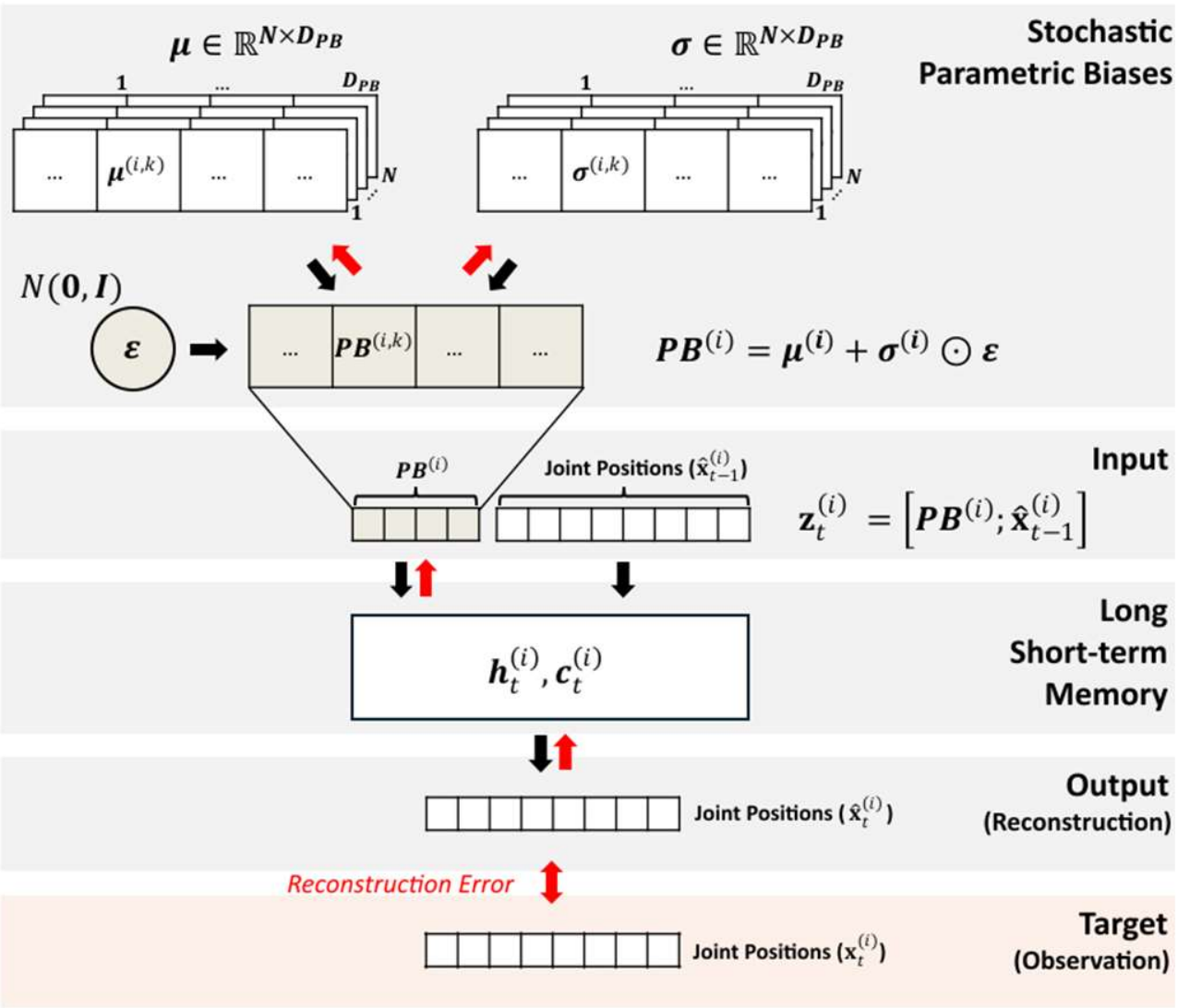

Fig. 1. Architecture of the proposed stochastic RNNPB (Recurrent Neural Network with Parametric Bias) model. The model consists of four main components: **Stochastic Parametric Biases (PB) Layer**: Introduces randomness by sampling a PB vector from a Gaussian distribution parameterized by mean $\boldsymbol{\mu}$ and standard deviation $\boldsymbol{\sigma}$ where $\epsilon$ is a random noise vector. **Input Layer**: Receives the input data at each time step, which is a concatenation of the PB value and the output from the previous time step. **Long Short-Term Memory (LSTM) Layer**: Processes sequential data by maintaining internal states that capture temporal dependencies. **Output Layer**: Generates the model's predictions at each time step. The black arrows indicate the top-down prediction of a sequence (reconstruction) and the red arrows denote the bottom-up recognition of a sequence (observation).

$\boldsymbol{\mu}^{(i)}$ and $\boldsymbol{\sigma}^{(i)}$ are the mean and standard deviation of the PB for sequence *i*, respectively. They are learnable parameters for each sequence, and each has a shape of ($N$, $D_{PB}$), where $N$ is the number of sequences in the training data, and $D_{PB}$ is the dimension of the PB vector. $\boldsymbol{\epsilon}$ is a random vector with elements sampled from a standard normal distribution, and $\odot$ denotes element-wise multiplication.

### *B. Training the Model*

The training process consists of iterations of the following steps: sampling PB from $N(\boldsymbol{\mu}, \boldsymbol{\sigma}^2)$, generating the output from the PB, computing the loss, and learning the parameters. In particular, the learning process can be understood within the Bayesian inference framework. During training, we aim to learn both the model parameters $\boldsymbol{\theta}$ (weights of the neural network) and the PB parameters ($\boldsymbol{\mu}$ and $\boldsymbol{\sigma}$). This process involves maximizing the Evidence Lower Bound (ELBO) [16], [34]. Specifically, for each sequence $\mathbf{x}$ in the training dataset, the goal is to infer the posterior distribution over the PB. Since the true posterior $p(\boldsymbol{PB} \mid \mathbf{x})$ is generally intractable, we use a variational distribution $q(\boldsymbol{PB})$ to approximate it. We parameterize $q(\boldsymbol{PB})$ as a Gaussian distribution with $\boldsymbol{\mu}$ and $\boldsymbol{\sigma}$. Then, our objective is to find the optimal $q(\boldsymbol{PB})$ and model parameters $\boldsymbol{\theta}$ by maximizing the ELBO (2), which indirectly minimizes the Kullback-Leibler (KL) divergence between $q(\boldsymbol{PB})$ and the true posterior.

$$\begin{aligned}\mathrm{ELBO}(\boldsymbol{\theta}, \boldsymbol{\mu}, \boldsymbol{\sigma}) = \; & \mathbb{E}_{q(\boldsymbol{PB})}[\log p(\mathbf{x} \mid \mathbf{PB}; \boldsymbol{\theta})] \\ & - \mathrm{KL}(q(\boldsymbol{PB}) \,||\, p(\boldsymbol{PB}))\end{aligned} \quad (2)$$

The first term of the ELBO is the expected log-likelihood of the data under the model and the approximate posterior. The second term is the KL divergence between the approximate posterior and the prior over PB. Maximizing the ELBO corresponds to finding model parameters ($\boldsymbol{\theta}$) and PB parameters ($\boldsymbol{\mu}$ and $\boldsymbol{\sigma}$) that best explain the training data while regularizing the complexity of the latent space. In practice, we minimize the negative ELBO, which corresponds to the loss function used during training (3).

$$\mathcal{L}_{total}(\boldsymbol{\theta}, \boldsymbol{\mu}, \boldsymbol{\sigma}) = \mathcal{L}_{recon} + \beta \times \mathcal{L}_{KL} \quad (3)$$

The loss function comprises a reconstruction term ($\mathcal{L}_{recon}$) and a regularization term ($\mathcal{L}_{KL}$). The reconstruction term corresponds to the negative expected log-likelihood, and it ensures that the model can accurately generate data from the PB and the model parameters. Under the Gaussian assumption, the reconstruction loss reduces to the Mean Squared Error (MSE) between the target and predicted data (4).

$$\mathcal{L}_{recon} = \frac{1}{N}\sum_{i=1}^{N}\sum_{t=1}^{T_i}\left\|\mathbf{x}_t^{(i)} - \hat{\mathbf{x}}_t^{(i)}\right\|^2 \quad (4)$$

where $T_i$ is the length of sequence $i$, $\mathrm{x}_t^{(i)}$ and $\hat{\mathrm{x}}_t^{(i)}$ are the target and predicted data at time $t$ respectively.

The regularization term involves the KL divergence, which measures how well the variational distribution matches the prior distribution over PB. Minimizing this term ensures that the approximate posterior distribution closely aligns with the assumed prior. This regularization helps maintain a simple and structured latent space [16], [30]. With the unit Gaussian prior assumption, the KL divergence for each sequence simplifies as follows (5).

$$\mathcal{L}_{reg} = \frac{1}{2N}\sum_{i=1}^{N}\sum_{j=1}^{D_{PB}}\left(\left(\mu_j^{(i)}\right)^2 + \sigma_j^{(i)2} - 1 - \ln \sigma_j^{(i)2}\right) \quad (5)$$

The beta term ($\beta$) in the loss function (3) serves as a weighting factor for the KL divergence, controlling the trade-off between reconstruction accuracy and regularization of the latent space. Previous studies [29] [30] have shown that a higher $\beta$ emphasizes the KLD term, leading to a more regularized and disentangled latent space but with reduced reconstruction accuracy. In the predictive coding framework, $\beta$ can be viewed as a parameter that adjusts the balance between reducing prediction errors (accuracy) and maintaining coherent internal models (complexity) [2].

### *C. Generating Sequences from Stochastic Parametric Biases*

The proposed model generates a sequence in an autoregressive manner, often referred to as closed-loop generation in previous studies [13], [17], [31]. At the onset of generation (*t=0*), PB is sampled, and the input to the model, as well as the initial states of the LSTM, are zeroed to mitigate the effect of different initial input values on the generation of different motions (i.e., $\hat{\mathbf{x}}_0^{(i)} = \mathbf{h}_0^{(i)} = \mathbf{c}_0^{(i)} = \mathbf{0}$). Consequently, only the PB values determine the type of sequences to be generated. Then, at each time step $t \geq 1$, the sampled PB and the model's output from the previous time step $t\text{-}1$ are fed into the LSTM layer (6). The output of the model is computed for each time step as in (7-8).

$$\mathbf{z}_t^{(i)} = \left[\boldsymbol{PB}^{(i)}; \hat{\mathbf{x}}_{t-1}^{(i)}\right] \quad (6)$$

$$\mathbf{h}_t^{(i)}, \mathbf{c}_t^{(i)} = \mathrm{LSTM}\left(\mathbf{z}_t^{(i)}, \mathbf{h}_{t-1}^{(i)}, \mathbf{c}_{t-1}^{(i)}\right) \quad (7)$$

$$\hat{\mathbf{x}}_t^{(i)} = \mathbf{W}_{\mathrm{out}}\mathbf{h}_t^{(i)} + \mathbf{b}_{\mathrm{out}} \quad (8)$$

where $\mathbf{W}_{\mathrm{out}}$ and $\mathbf{b}_{\mathrm{out}}$ are weights and biases in the output layer. Note that PB is sampled only at the initial time step rather than at every subsequent time step. This is to capture the sequence-level latent variables that represent the underlying structure of the sequence. Sampling the PB once ensures that the stochasticity remains consistent throughout the sequence, providing a stable and coherent influence on the model's dynamics and facilitating faster convergence.

### *D. Recognizing Sequences via Prediction Error Minimization*

The proposed model is able to recognize a sequence through prediction error minimization (PEM), which is one of the core ideas in predictive coding [1], [2]. Here, the term 'recognition' refers to posterior estimation during which the model updates the Gaussian parameters of the PBs for the given target sequence (observation). In standard VAE, recognition is implemented as a feedforward computation of the encoder network, which generates latent representations by using the observation as input. This process does not involve iterative optimization or PEM. In the proposed model, recognition is implemented as an iterative optimization of $\boldsymbol{\mu}$ and $\boldsymbol{\sigma}$ with PEM, in which they can be flexibly updated to align with the observation. This approach to recognition is similar to previous studies [8], [13], [31] and is rooted in Helmholtz's notion of perception as 'unconscious inference' [35].

To be more specific, the goal of recognition in the proposed model is to infer the PB values that best explain the observation. This process can be framed as performing approximate posterior estimation, similar to the training process. That is, we approximate the true posterior $p(\boldsymbol{PB} \mid \mathbf{x}_{obs})$ with a variational distribution $q(\boldsymbol{PB})$, and the goal is framed as finding $q(\boldsymbol{PB})$ that minimizes the KL divergence to the true posterior (9).

$$\min_{\boldsymbol{\mu},\boldsymbol{\sigma}} \mathrm{KL}(q(\boldsymbol{PB}) \mid\mid p(\boldsymbol{PB} \mid \mathbf{x}_{\mathrm{obs}})) \tag{9}$$

where $\mathbf{x}_{\mathrm{obs}}$ is an observed data. It should be noted that in (9), we focus on adjusting only $\boldsymbol{\mu}$ and $\boldsymbol{\sigma}$, not $\boldsymbol{\theta}$. In other words, the model parameters $\boldsymbol{\theta}$ are kept fixed during the recognition phase. Previous studies [15], [17], [18] have shown that this approach enables inference without compromising the model's long-term knowledge learned in $\boldsymbol{\theta}$. Additionally, learning the entire parameters during recognition could lead to overfitting to specific observations.

In practice, we simplify the optimization by neglecting the regularization term and only minimizing the reconstruction loss $\mathcal{L}_{\mathrm{recon}}^{\mathrm{obs}}$. Consequently, the recognition objective can be formulated as (10).

$$\hat{\boldsymbol{\mu}}, \hat{\boldsymbol{\sigma}} \cong \arg\min_{\boldsymbol{\mu},\boldsymbol{\sigma}} \mathcal{L}_{\mathrm{recon}}^{\mathrm{obs}} = \arg\min_{\boldsymbol{\mu},\boldsymbol{\sigma}} \sum_{t=1}^{T_{\mathrm{obs}}} \|\mathbf{x}_t - \hat{\mathbf{x}}_t\|^2 \tag{10}$$

where $T_{obs}$ is the length of the observation. By minimizing $\mathcal{L}_{\mathrm{recon}}^{\mathrm{obs}}$, the model is optimized so that the reconstruction becomes closer to the observation. This approach to recognition is referred to as "recognition by reconstruction" or "recognition by prediction" [8]. The core idea is that if the model can accurately reconstruct the observation, it must have captured the essential features and underlying structure of that observation. This method also aligns with the brain's mechanism of minimizing prediction errors to refine internal models [1], [2]. Note that this is an approximation to the full variational inference objective, where we neglect the regularization imposed by the prior $p(\boldsymbol{PB})$. The model focuses on precisely reconstructing the observed sequence by minimizing the reconstruction loss only during recognition. This is ideal for tasks requiring close matches, such as motion recognition.

To enhance the convergence speed and leverage the exploratory benefits of stochasticity in the proposed model, we implemented an early update mechanism in addition to the gradient-based update. The early update method involves directly updating $\boldsymbol{\mu}$ to the current sampled **PB** when the reconstruction loss is smaller than a threshold. That is, $\boldsymbol{\mu}^{(s+1)} = \boldsymbol{PB}^{(s)}\ if\ \mathcal{L}_{\mathrm{recon}}^{\mathrm{obs},(s)} \leq L_{MIN}$ where $s$ denotes iteration step during recognition. $L_{MIN}$ is a predefined threshold that is updated to reflect the minimum loss so far from $s=1$. Empirically, we observed that this early update elicited faster convergence during recognition, as it effectively balances exploration and exploitation.

## IV. Experiments

### A. Robotic Motion Dataset

In our experiment, we used the REBL (Robotic Emotional Body Language)-Pepper dataset [36]. It includes a collection of 36 hand-designed animations for the Pepper robot. These animations were crafted to express emotions through the robot's body gestures, eye LED patterns, and non-linguistic sounds. We used the augmented version included in the dataset, which was generated by mirroring each posture. As a result, a total number of 72 motion sequences were used in our experiment. Each motion consists of 17 joint angles expressed in radians. The joints are HeadPitch, HeadYaw, HipPitch, HipRoll, KneePitch, LElbowRoll, LElbowYaw, LHand, LShoulderPitch, LShoulderRoll, LWristYaw, RElbowRoll, RElbowYaw, RHand, RShoulderPitch, RShoulderRoll, and RWristYaw.

### B. Model Configurations

The proposed model consists of a stochastic PB layer with four neurons ($D_{PB}=4$), followed by an LSTM layer containing 256 hidden units. The LSTM layer is succeeded by a linear layer that produces the output joint angles. The input and output layers have 17 neurons, each representing the robot's joint angles in radians.

In the proposed model, the $\beta$ parameter plays a crucial role by weighting the KLD term in the loss function. To explore the impact of $\beta$ on the model's performance, we examined the model with different $\beta$ settings, as well as the model with fully deterministic dynamics (i.e., $\boldsymbol{PB}^{(i)} = \boldsymbol{\mu}^{(i)}$). A higher and lower value of $\beta$ are referred to as a strong and weak prior respectively, as the beta term indicates the amount of influence of the prior on the model training. Consequently, we analyzed the model's performance in the following four conditions:

- Stochastic model with Strong Prior ($\beta$ = 1e-3)
- Stochastic model with Weak Prior ($\beta$ = 1e-6)
- Stochastic model with Zero Prior ($\beta$ = 0)
- Deterministic model

Regarding the choice of $\beta$, we observed that models with $\beta$ greater than 1e-3 were unable to learn the training data accurately. The model was trained for 50,000 epochs using the Adam optimizer [37] with a learning rate of 0.001. Based on the unit Gaussian prior assumption, $\boldsymbol{\mu}$ and $\boldsymbol{\sigma}$ of the stochastic model were initialized to 0 and 1 respectively, during training.

### C. Tasks

We evaluated our model on two tasks: reconstruction and recognition. In the reconstruction task, the model generates motion sequences autoregressively from the learned PB values. The training data comprises 72 sequences, resulting in 72 pairs of learned $\boldsymbol{\mu}$ and $\boldsymbol{\sigma}$ values. For each pair (i.e., each motion sequence), we sampled 100 PB values. As a result, a total number of 7,200 sequences were generated for analysis. This extensive sampling allows us to assess the model's ability to reconstruct the motions with the learned PB distributions.

In the recognition task, we presented ten novel patterns to the model. The goal of the recognition task is to accurately reconstruct observation by updating the PB (i.e., recognition by reconstruction [8]). The novel patterns were generated by adding noise, scaling, and shifting to the principal components of the training data (see the code for details). Since the novel patterns were also the robot's joint angles, the recognition task can be seen as recognizing the robot's own action during kinesthetic teaching or learning from demonstration [23]. For each novel pattern, we conducted ten trials to assess the robustness of the recognition performance. In each trial, the optimization of $\boldsymbol{\mu}$ and $\boldsymbol{\sigma}$ was carried out for 100 iterations using the Adam optimizer [37] with a learning rate of 0.1.

The initialization of $\boldsymbol{\mu}$ and $\boldsymbol{\sigma}$ significantly impacts the convergence of the optimization process. We tested three initialization methods:

- Baseline ($\boldsymbol{\mu} = \mathbf{0}$, $\boldsymbol{\sigma} = \mathbf{I}$)
- Learned ($\boldsymbol{\mu}$ = one of the learned $\boldsymbol{\mu}$ values, $\boldsymbol{\sigma} = \mathbf{I}$)
- Random ($\boldsymbol{\mu}$ = one of 10 random $\boldsymbol{\mu}$ values, $\boldsymbol{\sigma} = \mathbf{I}$)

In the baseline condition, we initialized the $\boldsymbol{\mu}$ and $\boldsymbol{\sigma}$ to 0 and 1 respectively (i.e., unit Gaussian assumption). In the learned and random conditions, the model performed a pre-search phase in which it generated a set of outputs using different $\boldsymbol{\mu}$ values. The $\boldsymbol{\mu}$ values that resulted in the least reconstruction error (i.e., the output most similar to the observation) were chosen for the initialization. This technique is referred to as a "warm start", and it has been shown to improve optimization in previous studies [38], [39]. Note that $\boldsymbol{\sigma}$ was set to a very small positive value during the pre-search phase to minimize the effect of stochasticity in sampling PB. During the recognition phase, $\boldsymbol{\sigma}$ was initialized to one in all conditions, representing uncertainty at the beginning of recognition.

## V. Results

### A. Learning Stochastic Motion Representations

Fig. 2 depicts the probability density functions (PDFs) of PBs for three training sequences. As $\beta$ decreases, the PDFs become more spiky. This indicates that the model learns to assign lower variances to each sequence due to the lower regularization loss. In addition, the figure shows that the variance of the PB varies across sequences, demonstrating the model's capability to capture varying levels of uncertainty depending on the sequence.

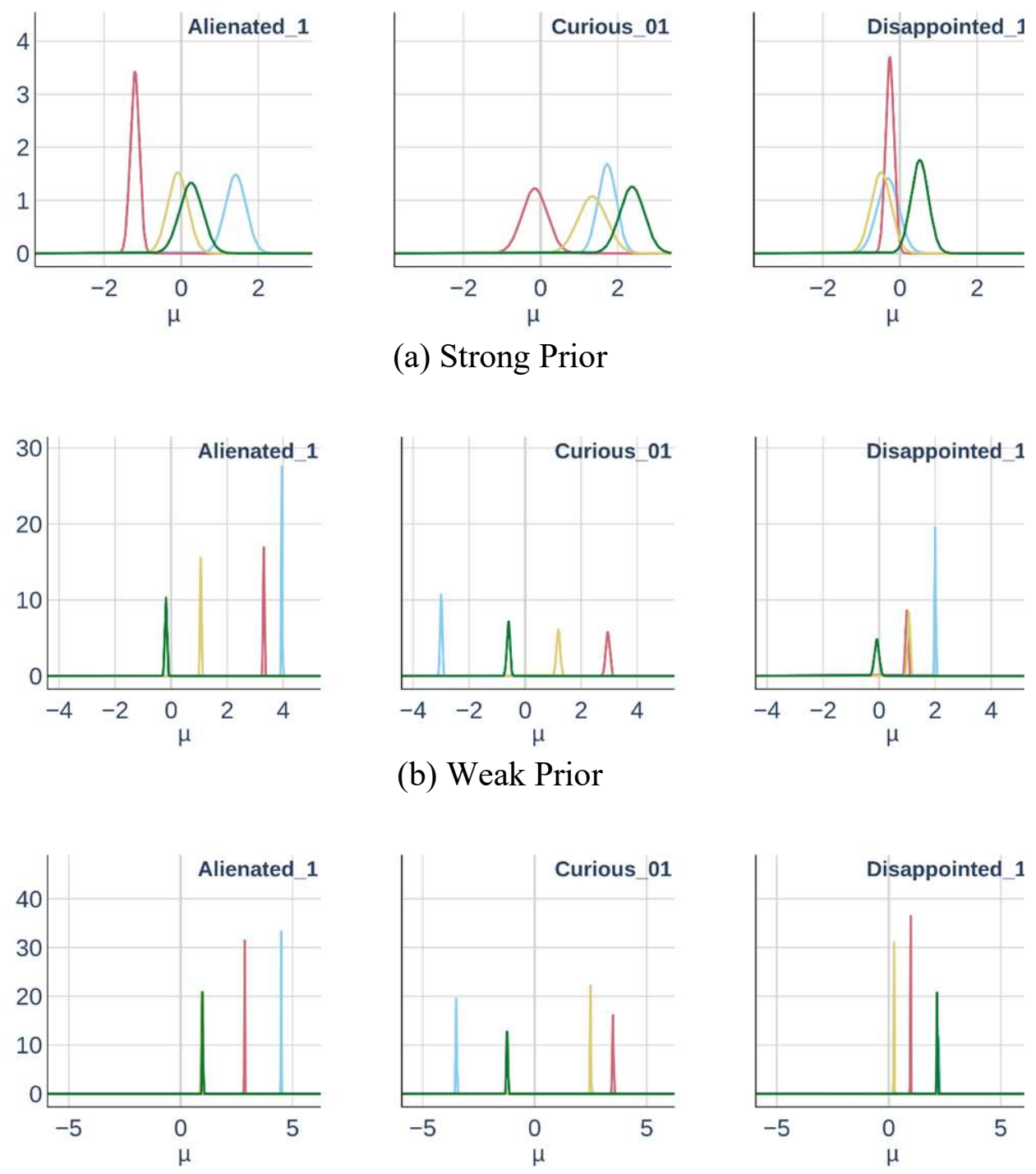

(a) Strong Prior

(b) Weak Prior

(c) Zero Prior

**Fig. 2.** Visualization of the probability distributions of the PB for three motion sequences across different model configurations. Each color corresponds to a different PB, represented by a Gaussian distribution with mean $\mu$ and standard deviation $\sigma$. The X-axis represents the PB values and the Y-axis shows the probability density. Broader distributions indicate greater uncertainty in the PB representation.

To understand the structure of the latent space, we conducted Principal Component Analysis (PCA) on the PB values. Fig. 3 presents PCA plots illustrating the distribution of PB values for different motion sequences across various model configurations. The stochastic model with a strong prior (high $\beta$) developed more dispersed representations than those with a weak or zero prior. This indicates a broader exploration of the latent space in the strong prior case. Conversely, in the zero prior case, the PB values for each motion cluster closely together, suggesting that the model learns more deterministic representations. This result demonstrates that as $\beta$ decreases, the model captures each motion sequence with increasing certainty and reduced variance due to the diminished effect of the KLD term.

### B. Generation of Sequences from the Stochastic PB

In the reconstruction task, we reconstruct the training sequences by sampling PB from the learned $\boldsymbol{\mu}$ and $\boldsymbol{\sigma}$ values. Table I summarizes the reconstruction loss for each model configuration, which indicates the average discrepancies between training and reconstructed sequences.

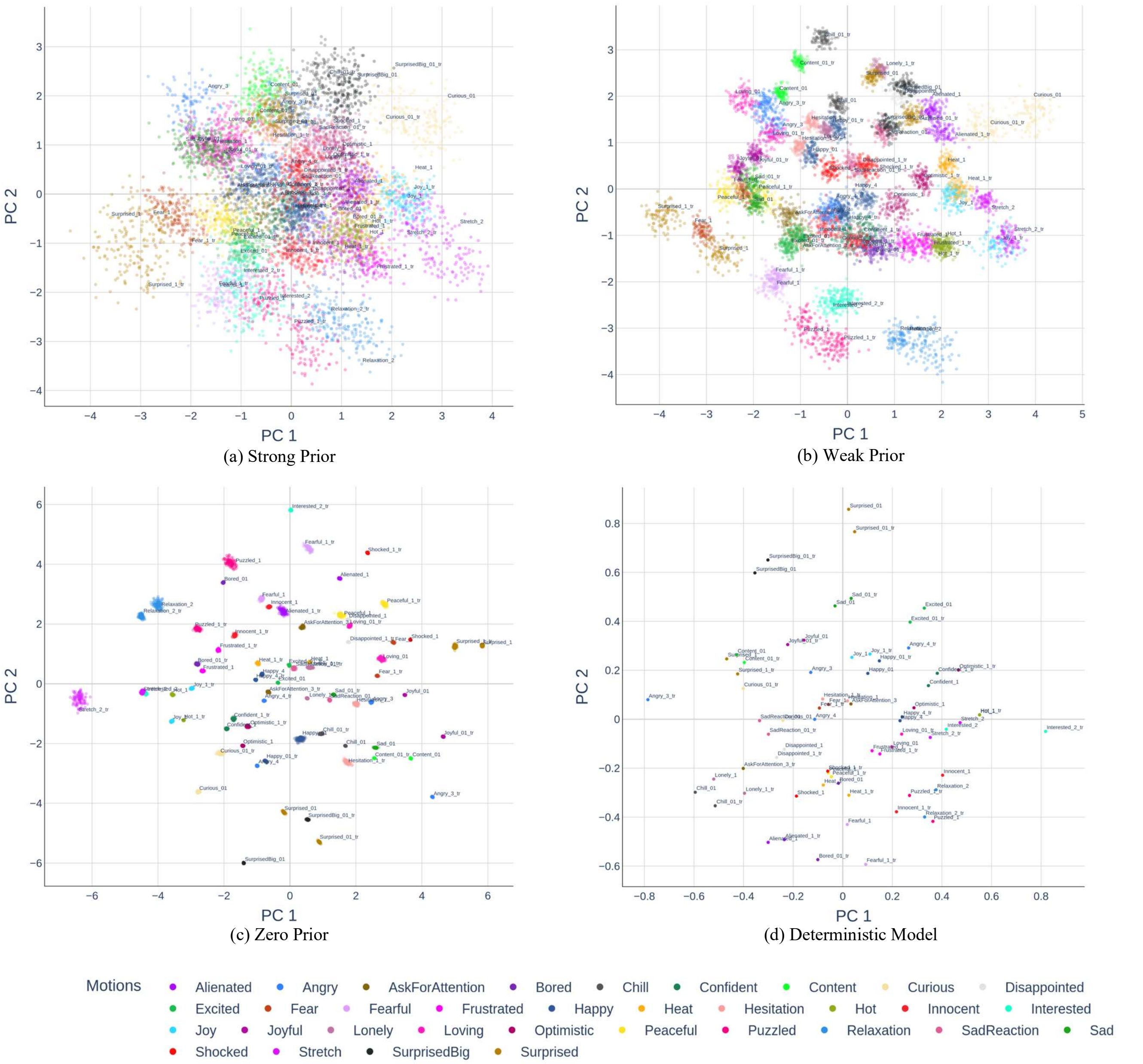

(a) Strong Prior

(b) Weak Prior

(c) Zero Prior

(d) Deterministic Model

**Fig. 3.** Principal Component Analysis (PCA) visualization of the PB distributions across different model configurations. The X and Y axes correspond to the first and second principal components, respectively. Each color corresponds to a distinct sequence from the training set. For each motion sequence, 100 samples of PB were generated by sampling from the learned $\boldsymbol{\mu}$ and $\boldsymbol{\sigma}$ values. The spread of points for each color indicates the uncertainty modeled by the stochastic PB. Since sampling is not applicable in the deterministic model (d), only the 72 training sequences are plotted.

In the stochastic models, the reconstruction loss decreases as $\beta$ decreases. This trend suggests that higher $\beta$, which enforces a stronger prior, leads to less accurate reconstructions. Interestingly, the reconstruction loss increases slightly with the size of the PB vectors in the stochastic models. On the other hand, the deterministic model exhibits a decrease in reconstruction loss as the PB size increases. This indicates a tendency toward overfitting in the deterministic model, as the model leverages more parameters to capture the training data. Overall, these results highlight the benefit of the stochastic model in that it may offer better generalization by avoiding overfitting regardless of hyperparameter settings.

We further examined the relationship between the latent space and the model output by systematically varying the PB values. Fig. 4 illustrates the landscape of PB in terms of the correlation between the training and reconstructed sequences.

The figure shows that the stochastic model develops a smoother latent space than the deterministic model. In the stochastic models, the correlation coefficient changes gradually as $\boldsymbol{\mu}$ changes regardless of $\beta$. This smoothness indicates that the models maintain a more continuous latent space around $\boldsymbol{\mu}$, allowing for more stable representations of the sequences. Moreover, the smoothness of the latent space also varies across different sequences. This indicates that the model self-organized the latent space, adapting its structure differently depending on the motion sequence. In contrast, in the deterministic model, the correlation coefficient changes drastically even with slight variations of $\boldsymbol{\mu}$. This suggests that the deterministic model is sensitive to minor perturbations, indicating the rugged latent space. This sensitivity may lead to poor generalization.

### *C. Recognition of Novel Sequences through Posterior Estimation of Stochastic PB*

We examined the model's recognition capability by presenting novel sequences (observations) and measuring how accurately the model could reconstruct these sequences by estimating the posterior. The recognition performance was measured in two terms: reconstruction loss and prediction error. The reconstruction loss refers to the discrepancy between the model output and the observed portion of the sequence (the first 80% of the sequence). This loss was used to update the PB values as described in Section III.D, and it indicates how well the model reconstructed the observation. The prediction error refers to the discrepancy between the model output and the remaining unobserved portion of the sequence. In other words, it indicates the model's ability to forecast or complete the motion sequence (i.e., to predict the latter 20% when given the first 80%). See Fig. 5 for an illustration of how the model performs recognition.

Table II presents the mean and standard deviation of the reconstruction loss and prediction error across different model configurations. The results demonstrated that the stochastic model outperformed the deterministic model in recognizing novel sequences. In particular, the stochastic models performed much better than the deterministic model in the baseline condition. A similar trend was also observed in prediction error, indicating better reconstruction performance led to improved forecast accuracy. In other words, by better capturing the first 80% of the sequences, the model could forecast the remaining 20% more accurately.

The results also showed that a warm start improved recognition performance. When the model was initialized with learned $\boldsymbol{\mu}$ values, both stochastic and deterministic models generally performed better than in the baseline conditions. Random search initialization also helped, but not as much as initialization with the learned $\boldsymbol{\mu}$ values. Notably, the deterministic model benefited the most from the warm start. Compared to the baseline condition, the deterministic model performed substantially better when initialized with learned $\boldsymbol{\mu}$. This implies the presence of a rugged loss landscape in the deterministic model, which often trapped the model in local minima during the baseline condition. The warm start helped the model start recognition at a favorable position in this landscape, enabling the model to avoid these minima. In contrast, the stochastic model demonstrated robust performance across all initialization conditions. The inherent stochasticity allowed the model to explore the loss landscape more broadly, reducing the likelihood of being trapped in suboptimal local minima. In summary, these results highlight that incorporating stochasticity into the model provides a significant advantage in handling uncertainty and improving recognition performance.

Fig. 6 visualizes the evolution of PB during the recognition process. The stochastic model explores a broader range of the latent space compared to the deterministic model, indicated by more dispersed PB values (green crosses). This stochastic exploration helps prevent the model from getting stuck in local

TABLE I

RECONSTRUCTION LOSS IN THE RECONSTRUCTION TASK

| $D_{PB}$ | *Stochastic Model* | | | *Deterministic Model* |
|---|---|---|---|---|
| | *Strong Prior* | *Weak Prior* | *Zero Prior* | |
| 1 | 0.055228 (0.072818) | 0.012470 (0.019954) | 0.003636 (0.003673) | 0.013471 (0.0173) |
| 2 | **0.003624** (0.009131) | 0.000035 (0.000017) | 0.000017 (0.00007) | 0.000079 (0.000051) |
| 4 | 0.005103 (0.014758) | **0.000032** (0.000012) | 0.000010 (0.000004) | 0.000018 (0.000008) |
| 8 | 0.006191 (0.016323) | 0.000043 (0.000631) | **0.000008** (0.000004) | 0.000010 (0.000004) |
| 16 | 0.006304 (0.016862) | 0.000372 (0.003565) | 0.000009 (0.000004) | 0.000011 (0.000005) |
| 32 | 0.007218 (0.019169) | 0.000115 (0.000087) | 0.000010 (0.000004) | 0.000009 (0.000004) |
| 64 | 0.008073 (0.021035) | 0.000294 (0.001455) | 0.000011 (0.000005) | 0.000009 (0.000004) |
| 128 | 0.011872 (0.02402) | 0.000521 (0.002697) | 0.000018 (0.000009) | **0.000008** (0.000003) |

TABLE II

RECONSTRUCTION LOSS AND PREDICTION ERROR IN THE RECOGNITION TASK

| *Warm Start* | *Stochastic Model* | | | *Deterministic Model* |
|---|---|---|---|---|
| | *Strong Prior* | *Weak Prior* | *Zero Prior* | |
| | *Reconstruction Loss* | | | |
| **Baseline** | 0.00225 (0.00054) | **0.00206** (0.00057) | 0.01439 (0.03887) | 0.13475 (0.05937) |
| **Learned** | 0.00227 (0.00054) | 0.00206 (0.00057) | **0.00201** (0.00054) | 0.00204 (0.00066) |
| **Random** | 0.00366 (0.00427) | 0.00212 (0.00056) | **0.00206** (0.00060) | 0.08153 (0.02994) |
| | *Prediction Error* | | | |
| **Baseline** | **0.00340** (0.00238) | **0.00340** (0.00293) | 0.00804 (0.01558) | 0.09514 (0.06043) |
| **Learned** | **0.00323** (0.00231) | 0.00334 (0.00272) | **0.00323** (0.00264) | 0.00420 (0.00344) |
| **Random** | 0.00385 (0.00250) | **0.00331** (0.00259) | 0.00332 (0.00272) | 0.07575 (0.05029) |

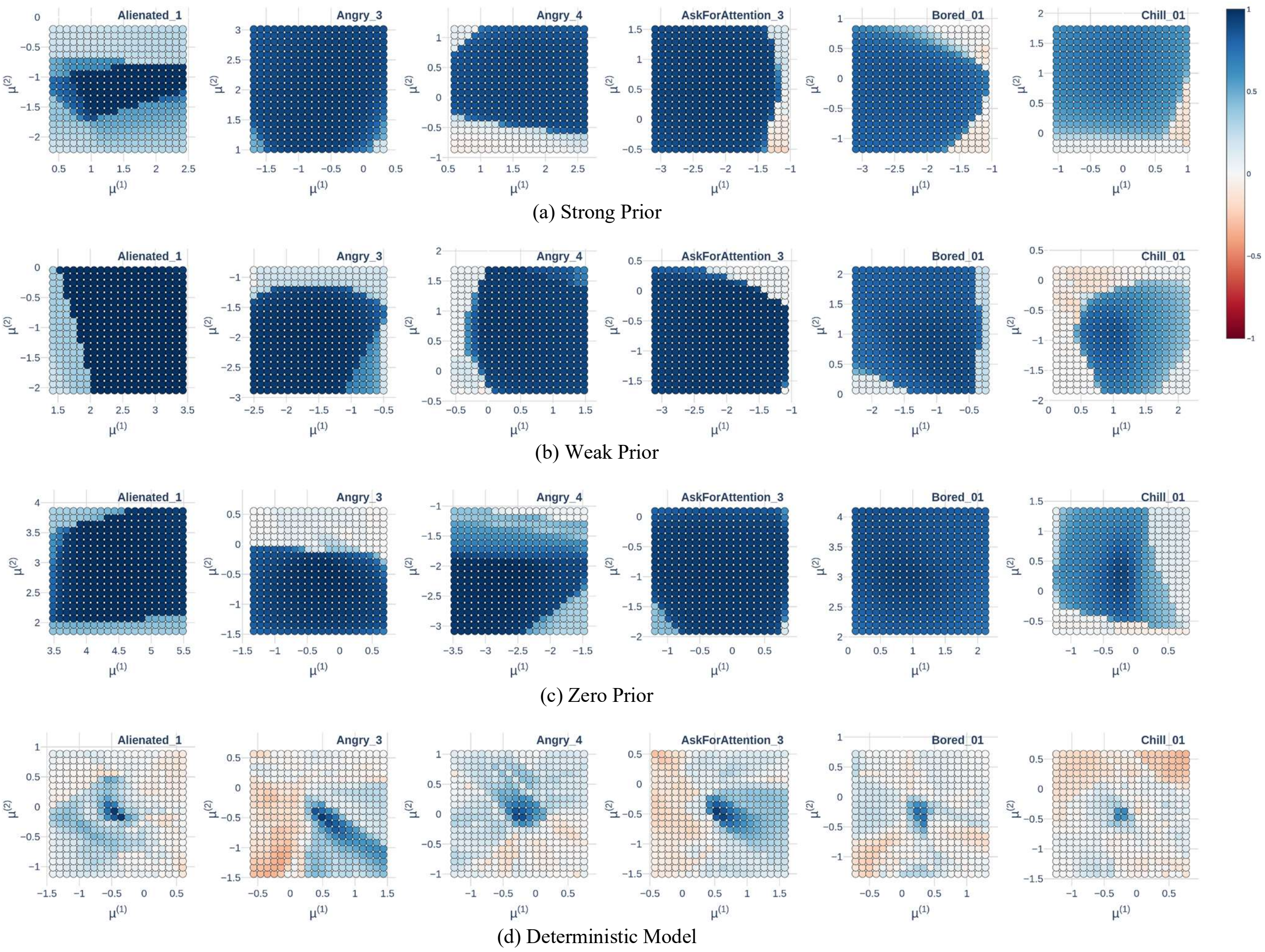

**Fig. 4.** Visualization of the latent space through correlation analysis. This figure illustrates how varying the $\boldsymbol{\mu}$ values of the PB affects the model's output. For each model, we visualize regions of the latent space for six target sequences, with each plot centered at the learned $\boldsymbol{\mu}^{(i,1)}$ and $\boldsymbol{\mu}^{(i,2)}$ of the sequence $\boldsymbol{i}$. The X and Y axes represent two varying $\boldsymbol{\mu}$ values, and we obtain 400 unique $\boldsymbol{\mu}$ pairs (20×20 grid). Each point represents a generated sequence using a specific PB vector defined by the $\boldsymbol{\mu}$ pair. Note that $\boldsymbol{\sigma}$ was set to zero to ensure that any variation in the generated sequences is solely due to changes in the $\boldsymbol{\mu}^{(i,1)}$ and $\boldsymbol{\mu}^{(i,2)}$. The color encodes the Pearson correlation coefficient between the generated sequence and the target sequence, indicating their similarities. The smoothness or abrupt changes of colors across the latent space provide insights into the landscape of the learned PB parameters.

minima, enhancing recognition outcomes. Moreover, the stochastic model exhibits different optimization paths in each trial due to its stochasticity. In contrast, the deterministic model shows the same narrower trajectories across different optimization trials. This illustrates the deterministic model's dependence on the initialization of PB, explaining the substantial benefits of a warm start for the deterministic model.

Additionally, we observed that the variance of PB changed over iterations during the recognition process. This indicates that the model actively adjusted the level of uncertainty in its beliefs based on the difference between the observation and reconstruction. Variance tended to decrease over iterations, although not monotonically. These results illustrate the model's ability to quantify and reduce uncertainty in response to the observation.

## V. Discussion

The experimental results have demonstrated several essential characteristics of the proposed model. In this section, we discuss these findings and the key characteristics of the model in detail, highlighting the benefits of the proposed model over the deterministic counterpart.

### *A. Bayesian Inference and Uncertainty Quantification with Stochastic Parametric Biases*

The proposed model learns PB as probability distributions parameterized by $\boldsymbol{\mu}$ and $\boldsymbol{\sigma}$, whereas previous deterministic

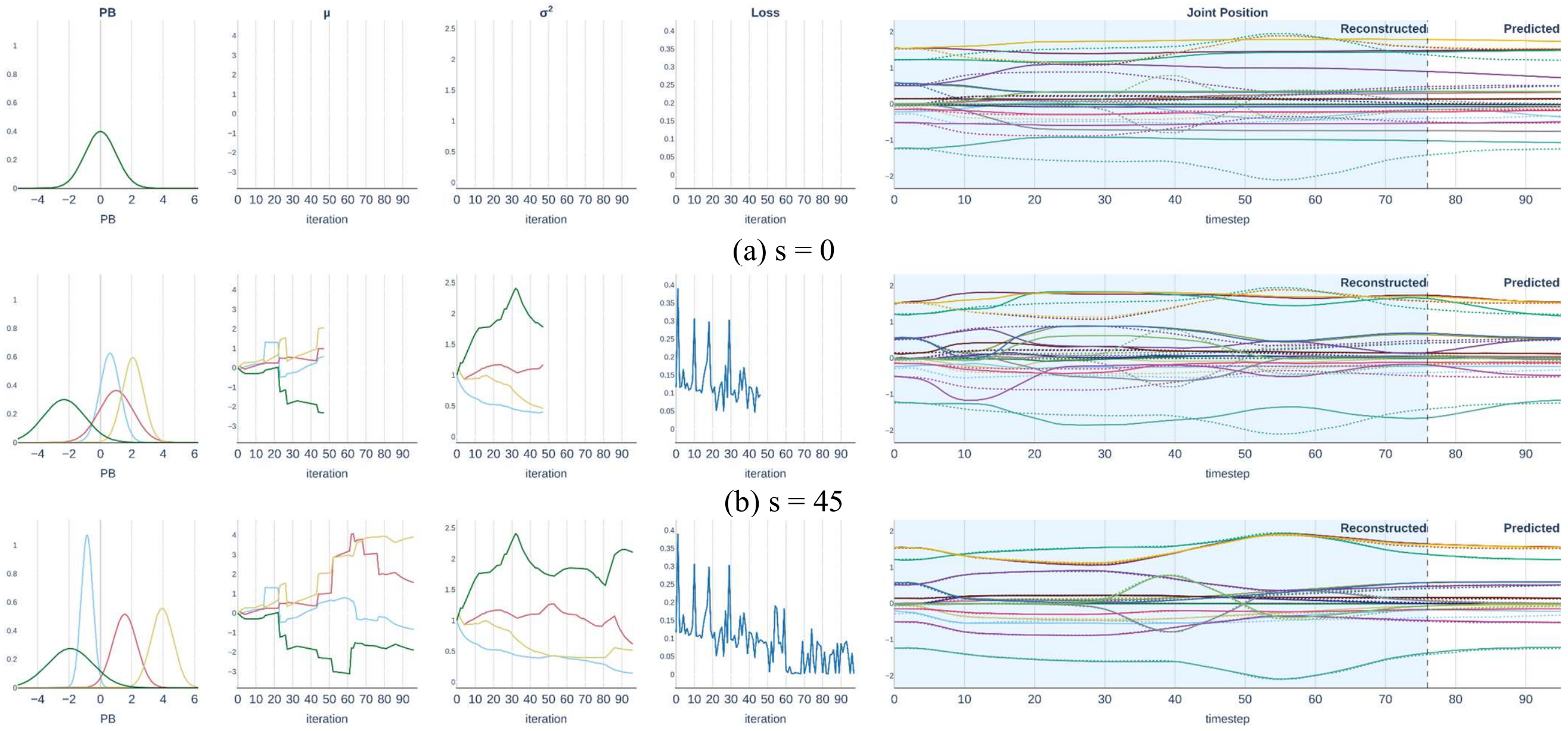

(a) s = 0

(b) s = 45

(c) s = 90

**Fig. 5.** Illustration of the recognition task at three optimization steps (s = 0, 45, 90). This figure demonstrates how the PB and outputs evolve during recognition. The first column shows the probability distributions of the PB and each color corresponds to a different PB. At s = 0, the PB is initialized as a unit Gaussian, indicating high uncertainty at the onset of recognition. As the optimization progresses, the PB distributions narrow, reflecting increased confidence and convergence towards specific $\boldsymbol{\mu}$ values. The second and third columns display the evolution of $\boldsymbol{\mu}$ and $\boldsymbol{\sigma}$ of the PB over iterations, representing the model's updated beliefs and the uncertainty in its belief, respectively. The fourth column shows the loss during recognition. The KL Divergence (orange line) is zero, indicating that optimization focuses solely on minimizing reconstruction error. A decreasing reconstruction loss (blue line) indicates the improved reconstruction accuracy. The fifth column compares the model output (solid line) to the target (dashed line) over time. As the model updates $\boldsymbol{\mu}$ and $\boldsymbol{\sigma}$ to minimize the reconstruction loss in the observation region (blue shaded area), its predictions in the prediction region become more accurate.

RNNPB models treat PB as fixed points. By learning a unique $\boldsymbol{\mu}$ and $\boldsymbol{\sigma}$ for each sequence in the training set, the proposed model encapsulates both the core representation and the uncertainty associated with the data. During recognition (i.e., inference), the model adjusts $\boldsymbol{\mu}$ and $\boldsymbol{\sigma}$ to minimize the discrepancy between reconstructions and observations. This is analogous to updating beliefs to minimize prediction error in predictive coding [1], [2] such that $\boldsymbol{\mu}$ represents the model's belief that explains the observation, and $\boldsymbol{\sigma}$ measures the uncertainty the model has in its belief $\boldsymbol{\mu}$.

Having probabilistic representations of PB endows the proposed model with several key features. First, the model can perform Bayesian inference using variational inference (VI). This is achieved by approximating the posterior with $\boldsymbol{\mu}$ and $\boldsymbol{\sigma}$ of PB, similar to VAE [16]. One notable difference between the proposed model and the VAE is in how the posterior estimation is computed after training. In general, after VAE models are trained, the posterior estimation (i.e., recognition) is done by a feedforward computation of the encoder network with the observation as input [16], [25]. This approach to posterior estimation can be referred to as sensory entrainment, in which the latent representation is driven by the forward computation of sensory inputs [8], [13].

In contrast, posterior estimation in our model is an iterative optimization process in which feedforward (reconstructing observation from the PB) and feedback (backpropagating reconstruction error) processes continuously interplay on the same network. This is achieved by utilizing the Mirror Neuron System (MNS)-like feature of RNNPB [18] which tightly intertwines generation and recognition of motion. In addition, our approach to posterior estimation is in line with previous studies [2], [5], in which perception is considered as an optimization that combines sensory input with prior expectation.

Second, the proposed model can represent uncertainty, which is valuable for tasks involving noisy or incomplete data [19], [20], [21], [34]. In the proposed model, representing uncertainty helps in both training and inference. During training, the stochasticity introduces variability, acting as a regularizer that prevents overfitting by utilizing the broader latent space. During recognition, the randomness allows the model to perform probabilistic inference, examining diverse solutions on the latent space during optimization. A higher $\boldsymbol{\sigma}$ means that the model is more open to exploring the latent space due to the variability in sampling PB. This is analogous to the precision of predictions (inverse of uncertainty) in a way that low precision (high uncertainty) leads the model to be more flexible in adjusting its beliefs [40]. In contrast, the deterministic model cannot generally perform probabilistic inference, potentially missing out on finding more suitable latent representations. Our finding supports previous studies that demonstrated the benefits of stochastic models in capturing complex data distributions and improving robustness to noise [16], [19], [20], [26], [30], [34].

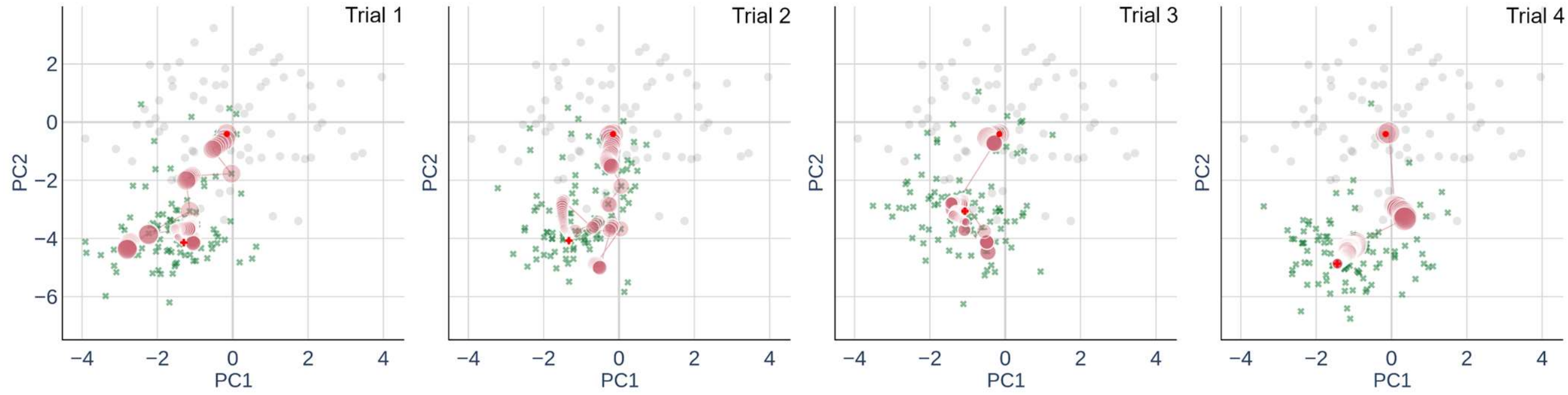

(a) Weak Prior

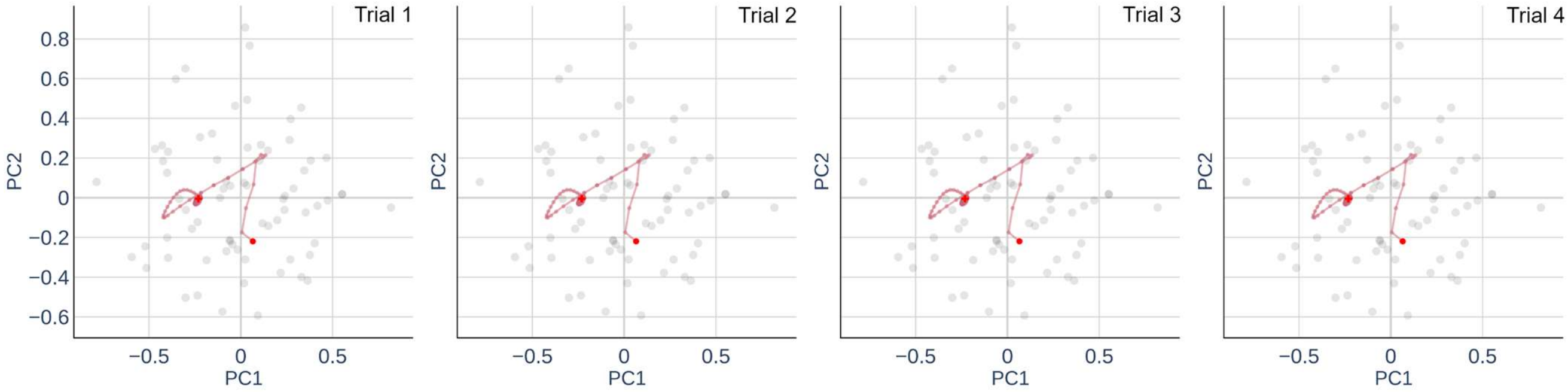

(b) Deterministic Model

**Fig. 6.** Evolution of the PB during the recognition task. Four recognition trials of the same novel sequence are shown for both stochastic and deterministic models. The evolution is projected onto the first two principal components of the PB. Red trajectories represent $\boldsymbol{\mu}$ values over iterations, with each red circle corresponding to an iteration during recognition. The size of each red circle represents the variance of the PB at that iteration, scaled relative to the initial unit variance for comparison. Larger circles indicate higher variance (greater uncertainty), while smaller circles indicate lower variance (increased certainty). A red point indicates the initial position, while a red cross indicates the terminal position. Green crosses depict the stochastic PB values sampled from $\boldsymbol{\mu}$ and $\boldsymbol{\sigma}$ at each iteration. Grey dots represent the learned $\boldsymbol{\mu}$ values of PB for the 72 training sequences, serving as reference locations in the PB space. Due to the stochasticity, the trajectories of $\boldsymbol{\mu}$ and PB values cover broader areas and differ in each trial in the stochastic model (a). In the deterministic model (b), the lack of stochasticity resulted in a narrow and identical path over different recognition trials.

Note that the stochastic PB illustrates uncertainty in the internal parameters ($\boldsymbol{\mu}$) that govern the model's behavior rather than in predictions themselves. Previous studies have incorporated uncertainty at different levels within neural network models. For instance, Blundell et al. [19] introduced probability distributions over the weights of neural network models. Uncertainty has also been incorporated in hidden units [13], [28] and outputs [40]. While these approaches offer the benefits of stochastic modeling, modeling uncertainty at these levels can be computationally intensive due to the large number of parameters involved. In contrast, we introduced uncertainty directly into PB, resulting in the model's predictions that are centered around internal beliefs $\boldsymbol{\mu}$ with uncertainty $\boldsymbol{\sigma}$. This is beneficial for tasks requiring adaptability and learning from limited data, such as in robotics. Moreover, modeling uncertainty in PB is less computationally expensive than modeling uncertainty at weights, hidden neurons, or outputs for every prediction.

### *B. Controlling the Level of Stochasticity with $\beta$*

We found that the choice of $\beta$ modulated the model's behavior. With a strong reliance on the KLD term (high $\beta$), the model developed a smoother latent space, generating more stable motions from the latent space, yet with lower fidelity. In contrast, a weak reliance on the KLD term (low $\beta$) resulted in the model behavior similar to that of the deterministic model. This finding aligns with previous studies [29], [30] that showed the trade-offs between reconstruction error and the quality of disentanglement in VAEs. Similarly, the $\beta$ term can also be understood in the predictive coding framework which posits that the brain balances the minimization of prediction errors with the complexity of its internal models [2].

We also found that setting $\beta$ to zero did not diminish the benefits of stochasticity. In the zero prior condition, the regularizing effect of the KLD term was removed, and the model focused solely on minimizing reconstruction error. Consequently, the learned $\boldsymbol{\sigma}$ values became very small, making the sampling process nearly deterministic. However, the zero prior model performed better than the deterministic model in both tasks. The latent space of the zero prior model was still smoother compared to the one of the deterministic model. This might be due to the unit Gaussian initialization of PB in training. As a result, the model still included stochastic elements from the sampling that regularized the latent space during the earlier stages of the training process. Nonetheless, setting $\beta$ to zero eventually removed the regularizing effect of the KLD term, resulting in lower performance than models with strong and weak priors. This suggests that including the KLD term with a positive $\beta$ value in the loss is crucial to ensure the model learns robust latent representations.

The stochasticity of the proposed model enhances the representational capacity. We observed that the stochastic model's representations were more widely distributed in the latent space, whereas the deterministic model's representations were closely clustered. Correlation analysis of the PB also suggests that the latent space of the deterministic model is rugged, whereas that of the stochastic model is smooth. This broader and smoother representation distribution in the stochastic model indicates a more comprehensive capture of the data variability, facilitating better generalization to unseen data. In contrast, the closely located representations in the deterministic model suggest that it is not effectively capturing the diversity of the data.

The advantages of having a smooth latent space have been highlighted in our tasks. The stable latent space enhances the model's robustness to perturbations. In the reconstruction task, small perturbations to $\boldsymbol{\mu}$ did not significantly alter the output in the stochastic model. In contrast, similar perturbations in the deterministic model led to drastic changes in the model output. The stable and consistent model behavior is crucial, particularly in contexts such as human-robot interaction or collaborative robotics when faced with noise or uncertainties in perception or internal representations.

The recognition task, especially with different initialization conditions, also highlighted the benefits of smooth latent space for optimization. The stochastic models outperformed the deterministic model for recognizing novel sequences across different initialization conditions. In contrast, the performance of the deterministic model heavily depended on the initialization method. This is because the deterministic model follows a single optimization path, so where to start optimization plays a crucial role in the task. The rugged latent space of the deterministic model makes it harder to generalize and explore different patterns, often leading to the model getting trapped in local minima. In other words, the deterministic model's latent space acted as an overly rigid prior, biasing the model's perception and causing it to ignore new sensory observations. This is parallel to the findings of [2], [41], in which too strong and rigid priors lead to a perceptual bias, hindering the brain from updating its predictions adequately in response to new sensory information.

### *C. Biologically-inspired Framework for Robotics and AI*

The proposed model is based on the fundamental principles of our brain—predictive coding [1], [2] and the Bayesian brain [3], [4]. By representing PB probabilistically, the proposed model incorporates uncertainty in its internal beliefs that govern its behavior. Moreover, the proposed model adjusts its beliefs and the level of uncertainty in its beliefs to refine its estimates in light of observed data.

The biologically inspired features of the proposed model offer several benefits. First, it can lead to more efficient and generalizable AI and robotic systems [6], [9]. Several studies have shown that biologically inspired architectures have been successfully applied in the field of machine learning, including computer vision applications [10], [11] and sequence modeling [14]. The proposed model's ability to quantify and adjust uncertainty through prediction error minimization can improve the performance of robotic systems that operate with noisy sensor and control signals.

Second, the proposed model is applicable for modeling differences in cognitive processing. For instance, according to [41], attenuated priors may result in a reduced capacity for generalization and more accurate perception in autistic individuals. Similarly, our experiments showed that reducing reliance on the prior resulted in worse generalization performance yet better reconstruction accuracy. Recently, this aspect has been examined in cognitive robotics to understand aberrant behaviors [40], [42], [43]. The capability to simulate different behavior by changing $\beta$ makes the proposed model a valuable tool in this field. Furthermore, the simplicity of the proposed model, compared to those with complex stochastic dynamics in the hidden states or weights, offers a straightforward and efficient framework for researchers, making analysis more accessible and interpretable.

## V. Conclusion

We introduced a novel stochastic RNNPB model that incorporates stochasticity into the parametric bias using the reparameterization trick. This approach allows the proposed model to learn probabilistic representations of multidimensional sequences, effectively capturing uncertainty and enabling variational inference through prediction error minimization. By aligning with predictive coding and the Bayesian brain hypothesis, our model offers a biologically inspired framework for sequence generation and recognition.

The proposed model was validated on a robotic motion dataset. The results revealed that the stochastic RNNPB model learned richer and more robust motion representations than its deterministic counterpart. In the latent space of the stochastic model, diverse robot motions were represented smoothly and continuously, enabling stable and robust motion generation and recognition. In contrast, the deterministic model learned point estimates for each sequence, resulting in a rugged latent space. As a result, the deterministic model was prone to overfitting and showed inferior performance compared to the stochastic model in the experiments.

Our approach provides a biologically inspired framework for modeling multidimensional sequences with stochasticity in machine-learning and robotics tasks. Furthermore, the proposed model could be used to simulate cognitive processes in cognitive and developmental robotics. However, several challenges remain for future research. First, the inference during recognition is computationally intensive due to the iterative optimization. It is worth studying how quickly the model can perform iterative optimization with limited computational resources. Also, extending the model to integrate multiple sensory modalities could enable it to capture more complex data representations.